\documentclass[a4paper,twoside]{article}

\usepackage{epsfig}
\usepackage{subfigure}
\usepackage{calc}
\usepackage{amssymb}
\usepackage{amstext}
\usepackage{amsmath}
\usepackage{amsthm}
\usepackage{multicol}
\usepackage{pslatex}
\usepackage{apalike}
\usepackage{SciTePress}
\usepackage[small]{caption}

\usepackage{algorithm}
\usepackage{algorithmic}

\begin{document}

\title{Multiclass Diffuse Interface Models for \\ Semi-Supervised Learning on Graphs}

\author{\authorname{Cristina Garcia-Cardona\sup{1}, Arjuna Flenner\sup{2} and Allon G.~Percus\sup{1}}
\affiliation{\sup{1}Institute of Mathematical Sciences, Claremont Graduate University, Claremont, CA 91711, USA}
\affiliation{\sup{2}Naval Air Warfare Center, Physics and Computational Sciences, China Lake, CA 93555, USA}
\email{cristina.garciacardona@cgu.edu, arjuna.flenner@navy.mil, allon.percus@cgu.edu}
}


\abstract{We present a graph-based variational algorithm for multiclass classification of high-dimensional data, motivated by total variation techniques. The energy functional is based on a diffuse interface model with a periodic potential. We augment the model by introducing an alternative measure of smoothness that preserves symmetry among the class labels. Through this modification of the standard Laplacian, we construct an efficient multiclass method that allows for sharp transitions between classes. The experimental results demonstrate that our approach is competitive with the state of the art among other graph-based algorithms.}

\onecolumn \maketitle \normalsize \vfill

\section{\uppercase{Introduction}}
\label{sec:introduction}

\noindent Many tasks in pattern recognition and machine learning rely on the ability to quantify local
similarities in data, and to infer meaningful global structure
from such local characteristics~\cite{coifman:lafon:lee}. In the
classification framework, the desired global structure is a descriptive
partition of the data into categories or classes. Many studies
have been devoted to the binary classification problems.  The multiple-class case, where the
data is partitioned into more than two clusters, is more
challenging.  One approach is to treat the problem as a series of
binary classification problems~\cite{allwein:schapire:singer}. In this
paper, we develop an alternative method, involving
a multiple-class extension of the diffuse interface
model introduced in~\cite{bertozzi:flenner}. 

The diffuse interface model by Bertozzi and Flenner combines methods for
diffusion on graphs with efficient partial differential equation techniques
to solve binary segmentation problems. As with other methods inspired by
physical phenomena~\cite{bertozzi:esedoglu:gillette,jung:kang:shen,li:kim}, it requires the minimization of an energy expression,
specifically the Ginzburg-Landau (GL) energy functional. The formulation
generalizes the GL functional to the case of functions defined on graphs,
and its minimization is related to the minimization of weighted graph
cuts~\cite{bertozzi:flenner}. In this sense, it parallels other
techniques based on inference on graphs via diffusion operators or
function estimation~\cite{coifman:lafon:lee,chung,zhou:scholkopf,szlam:maggioni:coifman,wang:jebara:chang,buhler:hein,szlam:bresson,hein:setzer}.

Multiclass segmentation methods that cast the problem as a series of
binary classification problems use a number of different strategies:
(i) deal directly with some binary coding or indicator for the
labels~\cite{dietterich:bakiri,wang:jebara:chang}, (ii) build a
hierarchy or combination of classifiers based on the one-vs-all approach
or on class rankings~\cite{hastie:tibshirani,har-peled:roth:zimak} or
(iii) apply a recursive partitioning scheme consisting of successively
subdividing clusters, until the desired number of classes is
reached~\cite{szlam:bresson,hein:setzer}. While there are advantages
to these approaches, such as possible robustness to mislabeled data,
there can be a considerable number of classifiers to compute, and
performance is affected by the number of classes to partition. 

In contrast, we propose an extension of the diffuse interface
model that obtains a simultaneous segmentation into multiple classes. 
The multiclass extension is built by modifying the GL energy functional to
remove the prejudicial effect that the order of the labelings,
given by integer values, has in the smoothing term of the original
binary diffuse interface model. A new term that promotes homogenization
in a multiclass setup is introduced. The expression penalizes data points
that are located close in the graph but are not assigned to the same
class. This penalty is applied {\em independently\/} of how
different the integer values are, representing the class labels.
In this way, the
characteristics of the multiclass classification task are incorporated
directly into the energy functional, with a measure of smoothness
independent of label order, allowing us to obtain high-quality results. Alternative multiclass methods minimize a Kullback-Leibler divergence function~\cite{subramanya:bilmes} or expressions involving the discrete Laplace operator on graphs~\cite{zhou:bousquet:lal,wang:jebara:chang}.

This paper is organized as follows. Section~\ref{sec:model} reviews the
diffuse interface model for binary classification, and describes its
application to
semi-supervised learning. Section~\ref{sec:multiclass} discusses our
proposed multiclass extension and the corresponding computational
algorithm.  Section~\ref{sec:results} presents results obtained with
our method. Finally,
section~\ref{sec:conclusion} draws conclusions and delineates future
work.

\section{\uppercase{Data Segmentation with the Ginzburg-Landau Model}} \label{sec:model}

\noindent The diffuse interface model~\cite{bertozzi:flenner} is based on a
continuous approach, using the Ginzburg-Landau (GL) energy functional to
measure the quality of data segmentation. A good segmentation is
characterized by a state with small energy. Let $u(\boldsymbol{x})$ be
a scalar field
defined over a space of arbitrary dimensionality, and representing the
state of the system.  The GL energy is written as the functional
\begin{equation}
E_{GL}(u) = \frac{\epsilon}{2} \int \! | \nabla u |^2 \; d\boldsymbol{x}
+ \frac{1}{\epsilon} \int \! F(u) \; d\boldsymbol{x},  \label{eq:GLf}
\end{equation}
\noindent with $\nabla$ denoting the spatial gradient operator,
$\epsilon > 0$ a real constant value, and $F$ a double well potential with minima at $\pm 1$:
\begin{equation}
	F(u) = \frac{1}{4} \left ( u^2 - 1 \right )^2. \label{eq:2pot}
\end{equation}

Segmentation requires minimizing the GL functional. The norm of the
gradient is a smoothing term that penalizes variations in the
field $u$. The potential term, on the other hand, compels $u$ to adopt
the discrete labels of $+1$ or $-1$, clustering the state of
the system around two classes. Jointly minimizing these two terms pushes
the system domain towards homogeneous regions with values close to the
minima of the double well potential, making the model appropriate for
binary segmentation. 

The smoothing term and potential term are in conflict at the interface
between the two regions, with the first term favoring a gradual
transition, and the second term penalizing deviations from the discrete
labels. A
compromise between these conflicting goals is established via the
constant $\epsilon$. A small value of $\epsilon$ denotes a small length
transition and a sharper interface, while a
large $\epsilon$ weights the gradient norm more, leading to a slower
transition. The result is a diffuse interface between regions, with
sharpness regulated by $\epsilon$. 

It can be shown that in the limit
$\epsilon \to 0$ this function approximates the total variation (TV)
formulation in the sense of functional ($\Gamma$)
convergence~\cite{kohn:sternberg},
producing piecewise constant solutions but with greater computational
efficiency than conventional TV minimization methods.
Thus, the diffuse interface model provides a framework
to compute piecewise constant functions with diffuse transitions,
approaching the ideal of the TV formulation, but with the
advantage that the smooth energy functional is more tractable
numerically and can be minimized by simple numerical methods such
as gradient descent.

The GL energy has been used to approximate the TV norm for image
segmentation~\cite{bertozzi:flenner} and image
inpainting~\cite{bertozzi:esedoglu:gillette,dobrosotskaya:bertozzi_inpainting}.
Furthermore, a calculus on graphs equivalent to TV has been introduced in~\cite{gilboa:osher,szlam:bresson}. 


\subsection*{Application of Diffuse Interface Models to Graphs}

An undirected, weighted neighborhood graph is used to represent the local
relationships in the data set. This is a common technique to segment classes that are not linearly separable.
In the $N$-neighborhood graph model, each vertex $z_i\in Z$ of the graph
corresponds to a data point with feature vector $\boldsymbol{x}_i$, while the
weight $w_{ij}$
is a measure of similarity between $z_i$ and $z_j$.
Moreover, it satisfies the symmetry property $w_{ij} = w_{ji}$. The
neighborhood is defined as the set of $N$ closest points in the feature
space. Accordingly, edges exist between each vertex and the vertices of
its $N$-nearest neighbors.  Following the approach of~\cite{bertozzi:flenner},
we calculate weights using the local
scaling of Zelnik-Manor and Perona~\cite{zelnik-manor:perona},
\begin{equation}
	w_{ij} = \exp \left ( - \frac{|| \boldsymbol{x}_i -
\boldsymbol{x}_j ||^2}{\tau(\boldsymbol{x}_i) \;
\tau(\boldsymbol{x}_j)} \right ). \label{eq:local_graph}
\end{equation}
Here, $\tau(\boldsymbol{x}_i) = ||\boldsymbol{x}_i -
\boldsymbol{x}^M_i||$ defines a local value for each $\boldsymbol{x}_i$,
where $\boldsymbol{x}^M_i$ is the position of the
$M$th closest data point to $\boldsymbol{x}_i$, and $M$ is a global
parameter.

It is convenient to express calculations on graphs via the graph
Laplacian matrix, denoted by $\boldsymbol{L}$.
The procedure we use to build the graph Laplacian is as follows.
\begin{enumerate}
\item Compute the similarity matrix $\boldsymbol{W}$ with components
$w_{ij}$ defined in (\ref{eq:local_graph}). As the neighborhood
relationship is not symmetric, the resulting matrix $\boldsymbol{W}$ is also not symmetric. Make it a symmetric matrix by connecting vertices $z_i$ and $z_j$ if $z_i$ is among the $N$-nearest neighbors of $z_j$ or if $z_j$ is among the $N$-nearest neighbors of $z_i$~\cite{luxburg}.
\item Define $\boldsymbol{D}$ as a diagonal matrix whose $i$th diagonal element represents the degree of the vertex $z_i$, evaluated as
\begin{equation}
  d_i = \sum_{j} w_{ij}.
\end{equation} 
\item Calculate the graph Laplacian: $\boldsymbol{L} = \boldsymbol{D} - \boldsymbol{W}$.
\end{enumerate}
Generally, the graph Laplacian is normalized to guarantee spectral
convergence in the limit of large sample size~\cite{luxburg}.
The symmetric normalized graph Laplacian $\boldsymbol{L_s}$ is defined
as
\begin{equation}
	\boldsymbol{L_s} = \boldsymbol{D}^{-1/2} \; \boldsymbol{L} \;
\boldsymbol{D}^{-1/2} = \boldsymbol{I} - \boldsymbol{D}^{-1/2} \;
\boldsymbol{W} \; \boldsymbol{D}^{-1/2}.
\label{eq:Ls}
\end{equation}

Data segmentation can now be carried out through a graph-based formulation of the GL energy. To implement this task, a fidelity term is added to the functional as initially suggested in~\cite{dobrosotskaya:bertozzi}. This enables the specification of a priori information in the system, for example the known labels of certain points in the data set. This kind of setup is called semi-supervised learning (SSL). The discrete GL energy for SSL on graphs can be written as~\cite{bertozzi:flenner}:
\begin{eqnarray}
\label{eqn:graphLaplacian}
E_{GL_{\mathrm{SSL}}}(\boldsymbol{u}) & = & \frac{\epsilon}{2} \langle
\boldsymbol{u}, \boldsymbol{L_s} \boldsymbol{u} \rangle + \frac{1}{\epsilon}
\sum_{z_i \in Z} F(u(z_i)) \nonumber \\ 
& &
+ \sum_{z_i \in Z} \frac{\lambda(z_i)}{2} \; \left ( u(z_i) - u_0(z_i) \right )^2
\end{eqnarray} 
\noindent In the discrete formulation, $\boldsymbol{u}$ is a
vector whose component $u(z_i)$ represents the state of the vertex
$z_i$,
$\epsilon > 0$ is a real constant characterizing the smoothness of the
transition between classes, and $\lambda(z_i)$ is a fidelity weight
taking value $\lambda > 0$ if the label $u_0(z_i)$ (i.e. class) of the
data point associated with vertex $z_i$ is known beforehand, or
$\lambda(z_i)=0$ if
it is not known (semi-supervised).

Equation (\ref{eqn:graphLaplacian}) may be understood as an example of
the more general form of an energy functional for data classification,
\begin{eqnarray}
E(\boldsymbol{u}) =  || \boldsymbol{u} ||_{a} + \frac{\lambda}{2} ||
\boldsymbol{u} - \boldsymbol{f} ||_{b}^{p},
\label{eqn:basic}
\end{eqnarray}
where the norm $||u||_{a}$ is a regularization term and $||u - f||_{b}$
is a fidelity term.  The choice of the regularization norm
$||\cdot||_{a}$ has non-trivial consequences  in the final
classification accuracy.  Attractive qualities of the norm $|| \cdot
||_{a}$ include allowing classes to be close in a metric space, and
obtain segmentations for nonlinearly separable data.  Both of these
goals are addressed using the GL energy functional for SSL.

Minimizing the functional simulates
a diffusion process on the graph. The information of the few labels
known is propagated through the discrete structure by means of the
smoothing term, while the potential term clusters the vertices around
the states $\pm 1$ and the fidelity term enforces the known labels. The energy minimization process itself attempts to reduce the interface regions.
Note that in the absence of the
fidelity term, the process could lead to a trivial
steady-state solution of the diffusion equation, with all data points
assigned the same label.

The final state $u(z_i)$ of each vertex is obtained by thresholding, and the resulting homogeneous regions with labels of $+1$ and $-1$ constitute the two-class data segmentation.

\section{\uppercase{Multiclass Extension}}
\label{sec:multiclass}

\noindent The double-well potential in the diffuse interface model for SSL flows the state of the system towards two
definite labels. Multiple-class segmentation requires a more general potential function $F(u)$ that allows
clusters around more than two labels. For this purpose, we use the
periodic-well potential suggested by Li and Kim~\cite{li:kim},
\begin{equation}
F( u ) = \frac{1}{2} \, \{ u \}^2 \, (\{ u \} - 1)^2, \label{eq:well_ext}
\end{equation}
where $\{ u \}$ denotes the fractional part of $u$,
\begin{equation}
\{ u \} = u - \lfloor u \rfloor, \label{eq:multiphase}
\end{equation}
\noindent and $ \lfloor u \rfloor$ is the largest integer not greater than $u$.

This periodic potential well promotes a multiclass solution, but the
graph Laplacian term in Equation (\ref{eqn:graphLaplacian}) also requires modification for effective calculations due to the fixed ordering of class labels in the multiple class
setting. The graph Laplacian term penalizes large changes in the spatial
distribution of the system state more than smaller gradual
changes. In a multiclass framework, this implies that the penalty for
two spatially contiguous classes with different labels may vary
according to the (arbitrary) ordering of the labels.

\begin{figure}[htb]
\begin{center}
\framebox{\scalebox{0.1}{\includegraphics{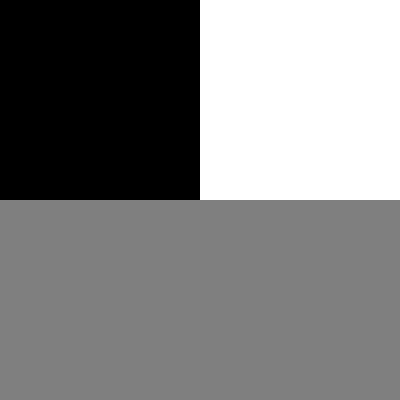}}}
\end{center}
\caption{Three class segmentation. Black: class 0. Gray: class 1. White: class 2.}
\label{fig:why_multiclass}
\end{figure}

This phenomenon is shown in
Figure~\ref{fig:why_multiclass}. Suppose that the goal is to segment the
image into three classes: class 0 composed by the black region, class 1
composed by the gray region and class 2 composed by the white region. It
is clear that the horizontal interfaces comprise a jump of size 1
(analogous to a two class segmentation) while the vertical interface
implies a jump of size 2. Accordingly, the smoothing term will assign a
higher cost to the vertical interface, even though from the point of
view of the classification, there is no specific reason for this.  In
this example, the problem cannot be solved with a different label
assignment. There will always be an interface with higher costs than
others independent of the integer values used.

Thus, the multiclass approach breaks the symmetry among classes,
influencing the diffuse interface evolution in an undesirable
manner. Eliminating this inconvenience requires restoring the symmetry,
so that the difference between two classes is always the same,
regardless of their labels. This objective is achieved by
introducing a new class difference measure. 

\subsection{Generalized Difference Function}

The 
final class labels are determined by thresholding each vertex $u(z_i)$, with the label $y_i$
set to the nearest integer:
\begin{equation}
y_i = \left \lfloor u(z_i) + \frac{1}{2} \right \rfloor.
\end{equation}

\begin{figure}
\begin{center}
\begin{picture}(150,70)(0,0)
\put(60,40){\circle*{5}}
\put(60,40){\line(-1,-1){40}}
\put(60,40){\line(0,-1){40}}
\put(60,40){\line(1,-1){40}}

\put(101,-1){\circle{2}}
\put(60,-1){\circle{2}}
\put(19,-1){\circle{2}}

\put(68,38){\mbox{Half-integer}}
\put(110,-5){\mbox{Integer}}

\put(55,45){\line(-1,-1){25}}
\put(50,50){\line(1,-1){10}}
\put(25,25){\line(1,-1){10}}
\put(25,38){\mbox{$\hat{r}(x)$}}
\end{picture}
\end{center}
\caption{Schematic interpretation of generalized difference:
$\hat{r}(x)$ measures distance to nearest half-integer, and
$\rho$ then corresponds to distance on tree.}
\label{fig:r_hat}
\end{figure}
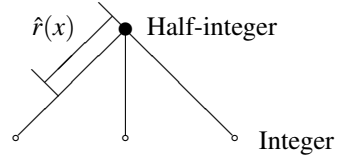

The boundaries between classes then occur at half-integer values corresponding to the unstable equilibrium states of the potential well.
Define the function $\hat{r}(x)$ to represent the distance to
the nearest half-integer:
\begin{equation}
\hat{r}(x) = \left | \frac{1}{2} - \{ x \} \right |. \label{eq:r_hat}
\end{equation}
A schematic of $\hat{r}(x)$ is depicted in Figure~\ref{fig:r_hat}. The $\hat{r}(x)$ function is used to define a generalized difference function between
classes that restores symmetry in the energy functional. Define the generalized difference function
$\rho$ as:

\begin{equation}
\rho(u(z_i),u(z_j)) = 
\left \{
\begin{array}{lll}
\hat{r}(u(z_i)) + \hat{r}(u(z_j)) & \ & y_i \neq y_j \\
& & \\
\left|\hat{r}(u(z_i)) - \hat{r}(u(z_j))\right| & & y_i = y_j
\end{array}
\right .
\end{equation}


Thus, if the vertices are in different classes, the difference
$\hat{r}(x)$ between each state's value and the nearest half-integer is added,
whereas if they are in the same
class, these differences are subtracted. The function $\rho(x,y)$ corresponds to
the tree distance (see Fig.~\ref{fig:r_hat}).  Strictly speaking,
$\rho$ is not a metric since it does not satisfy $\rho(x,y) =
0 \Rightarrow  x = y$.  Nevertheless, 
the cost of interfaces between classes becomes the same regardless of class
labeling when this generalized distance function is implemented.

The GL energy functional for SSL, using the
new generalized difference function $\rho$, is expressed as
\begin{eqnarray}
E_{MGL_{\mathrm{SSL}}}(\boldsymbol{u}) & = & \frac{\epsilon}{2}
\sum_{z_i \in Z} \sum_{z_j \in Z} \frac{w_{ij}}{\sqrt{d_i d_j}} \,
\left[\rho(u(z_i),u(z_j))\,\right]^2 \nonumber \\
& & + \frac{1}{2 \epsilon}\sum_{z_i \in Z} \{ u(z_i) \}^2 \, ( \{ u(z_i) \} - 1 )^2 \nonumber \\
& & + \sum_{z_i \in Z} \frac{\lambda(z_i)}{2} \; \left ( u(z_i) -
u_0(z_i) \right )^2.  \label{eq:multiclass_model}
\end{eqnarray}
Note
that $\rho$ could also be used in the fidelity term, but for simplicity this modification is not included.  In practice,
this has little effect on the results.

\subsection{Computational Algorithm}

The GL energy functional given by (\ref{eq:multiclass_model}) may be
minimized iteratively, using gradient descent:
\begin{equation}
u_i^{m+1} = u_i^{m} - dt \, \left [\frac{\delta
E_{MGL_{\mathrm{SSL}}}}{\delta u_i} \right ],
\end{equation}
where $u_i$ is a shorthand for $u(z_i)$, $dt$ represents the time step and the gradient direction is given by:

\begin{equation}
\frac{\delta E_{MGL_{\mathrm{SSL}}}}{\delta u_i} =  
\epsilon G(u_i^m) +   \frac{1}{\epsilon} F'(u_i^m) + \lambda_i \left ( u_i^m - {u_i}_0 \right ) 
\end{equation}

\begin{equation}
G(u_i^m) =  \sum_j \frac{w_{ij}}{\sqrt{d_i d_j}} \left [ \hat{r}(u_i^m) \pm \hat{r}(u_j^m) \right ] \hat{r}'(u_i^m) \label{eq:G} 
\end{equation}

\begin{equation}
F'(u_i^m) = 2 \; \{ u_i^m \} ^3 - 3 \; \{ u_i^m \} ^2 + \{ u_i^m \}
\end{equation}

\begin{algorithm*}[ht]
\caption{Calculate $\boldsymbol{u}$} \label{algo:iter}
\begin{algorithmic} 
\REQUIRE $\epsilon > 0, dt > 0, m_{\mathrm{max}} > 0, K \mathrm{~given}$
\ENSURE $\mathrm{out} = \boldsymbol{u}^{m_{\mathrm{max}}}$
\STATE $\boldsymbol{u}^0 \leftarrow rand((0,K))-\frac{1}{2}, \ m \leftarrow 0$
\FOR{$m < m_{\mathrm{max}}$}
\STATE $i \leftarrow 0$
\FOR{$i < n$}
\STATE $u_i^{m+1} \leftarrow u_i^m - dt \left ( \epsilon \: G(u_i^m)  +
\frac{1}{\epsilon} \: F'(u_i^m) + \lambda_i \left ( u_i^m - {u_i}_0 \right ) \right )$
\IF{$\mathrm{Label}(u_i^{m+1}) \neq  \mathrm{Label}(u_i^{m})$}
\STATE $(v_i)_k \leftarrow k + \{ u_i ^{m+1}\}$
\STATE $u_i^{m+1} \leftarrow (v_i)_k \mathrm{~where~} k=\arg\min_{\; 0
\leq k < K} \; \sum_{j } \frac{w_{ij}}{\sqrt{d_i d_j}} \,
\left[\rho((v_i)_k,u_j)\,\right]^2$
\ENDIF
\STATE $ i \leftarrow i + 1$
\ENDFOR
\STATE $m \leftarrow m + 1$
\ENDFOR
\end{algorithmic}
\end{algorithm*}

The gradient of the generalized difference function $\rho$ is not defined at half integer values.  Hence, we
modify the method using a greedy strategy: after detecting that a
vertex changes class, the new class that minimizes the
smoothing term is selected, and the fractional part of the state
computed by the gradient descent update is preserved. Consequently, the
new state of vertex $i$ is the result of gradient descent,
but if this causes a change in class, then a new state is determined.

Specifically, let
$k$ represent an integer in the range of the problem, i.e. $ k \in [0,
K-1]$, where $K$ is the number of classes in the problem. Given the
fractional part $\{u\}$ resulting from the gradient descent update,
define $(v_i)_k = k + \{u_i\}$.  Find the integer $k$ that minimizes
$\sum_{j} \frac{w_{ij}}{\sqrt{d_i d_j}} \,
\left[\rho((v_i)_k,u_j)\,\right]^2$, the smoothing term in the energy functional, and use $(v_i)_k$ as
the new vertex state.
A summary of the procedure is shown in Algorithm~\ref{algo:iter}
with $m_{\mathrm{max}}$ denoting the maximum number of iterations.

\section{\uppercase{Results}} 
\label{sec:results}

\noindent The performance of the multiclass diffuse interface model
is evaluated using a number of data sets from the literature, with differing
characteristics.  Data and image segmentation problems are considered on synthetic and real data sets.

\subsection{Synthetic Data}

A synthetic three-class segmentation problem is constructed following an
analogous procedure used in~\cite{buhler:hein} for ``two moon'' binary
classification, using three half circles (``three moons''). The half
circles are generated in
$\mathbb{R}^2$. The two top circles have radius $1$ and are centered
at $(0, 0)$ and $(3, 0)$. The bottom half circle has radius $1.5$ and
is centered at $(1.5, 0.4)$. We sample 1500 data points (500 from each of these
half circles) and embed them in $\mathbb{R}^{100}$. The
embedding is completed by adding Gaussian noise with $\sigma^2= 0.02$
to {\em each\/} of the 100 components for each data point. The
dimensionality of the data set, together with the noise,
make this a nontrivial problem. 

\begin{figure*}[tb]
\centerline{
\subfigure{\scalebox{0.3}{\includegraphics{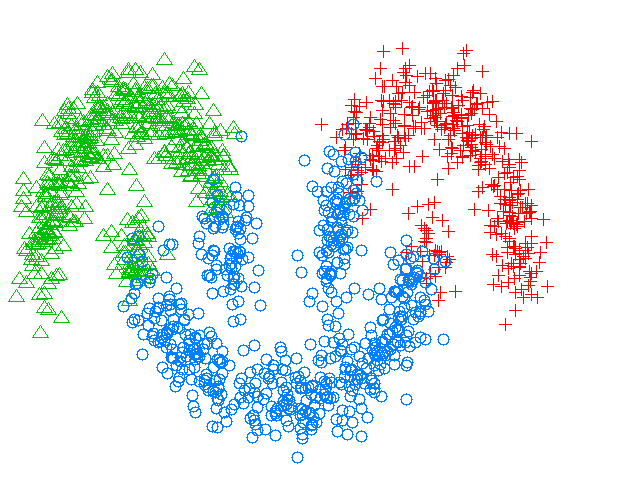}}}
\hfil
\subfigure{\scalebox{0.3}{\includegraphics{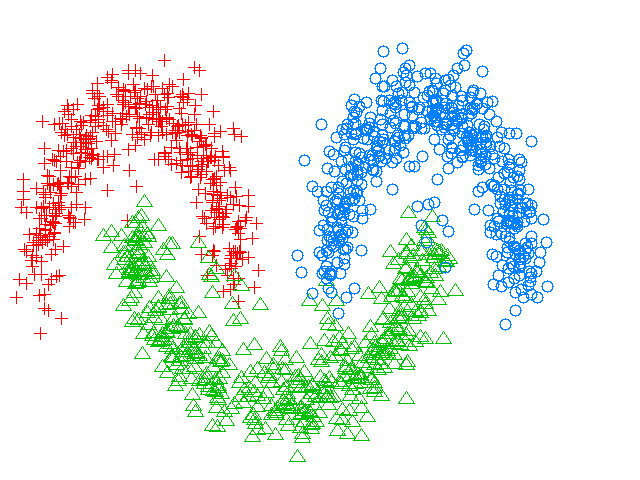}}}
}
\caption{Three-class segmentation. Left: spectral clustering. Right: multiclass GL (adaptive $\epsilon$).}
\label{fig:3moon}
\end{figure*}

The difficulty of the problem is illustrated in Figure~\ref{fig:3moon}, where we use both spectral clustering decomposition and the multiclass GL method. The same graph structure is used for both methods. The symmetric graph Laplacian is
computed based on edge weights given by (\ref{eq:local_graph}),
using $N = 10$ nearest neighbors and local scaling based on
the $M = 10$ closest point. The spectral clustering results are
obtained by
applying a $k$-means algorithm to the first $3$ eigenvectors of the
symmetric graph Laplacian. The average error obtained, over 100
executions of spectral clustering, is 20\% ($\pm 0.6\%$). The figure
displays the best result obtained, corresponding to an error of $18.67\%$.  

The multiclass GL method was implemented with the following parameters:
interface scale $\epsilon = 1$, step size $dt = 0.01$ and number of
iterations $m_{\mathrm{max}} = 800$.  The fidelity term is determined by
labeling 25 points randomly selected from each class (5\% of all
points), and setting the fidelity weight to $\lambda = 30$ for those
points.  Several runs of the procedure are performed to isolate effects from the random
initialization and the arbitrary selection of fidelity points. The average error obtained, over 100 runs with four
different fidelity sets, is 5.2\% ($\pm 1.01\%$). In general terms, the
system evolves from an initially inhomogeneous state, rapidly developing
small islands around fidelity points that become seeds for homogeneous
regions and progressing to a configuration of classes forming nearly
uniform clusters.

The multiclass results were further improved by incrementally decreasing $\epsilon$ to allow sharper transitions between states as in~\cite{bertozzi:flenner}. With this approach, the average error
obtained over 100 runs is reduced to 2.6\% ($\pm 0.3\%$). The best
result obtained in these runs is displayed in Figure~\ref{fig:3moon} and
corresponds to an average error of 2.13\%. In these runs, $\epsilon$ is
reduced from $\epsilon_0 = 2$ to $\epsilon_f = 0.1$ in decrements of
10\%, with $40$ iterations performed per step. The average computing
time per run in this adaptive technique is 1.53s in an Intel Quad-Core
@ 2.4 GHz, without any parallel processing. 

For comparison, we note the results from the literature for the
simpler two moon problem ($\mathbb{R}^{100}$, $\sigma^2= 0.02$ noise).
The best errors reported include: 6\% for p-Laplacian~\cite{buhler:hein},
4.6\% for ratio-minimization relaxed Cheeger cut~\cite{szlam:bresson},
and 2.3\% for binary GL~\cite{bertozzi:flenner}. While these are not SSL methods the last of these does involve other prior information in the form of a mass balance constraint.
It can be seen that both of our procedures, fixed and adaptive
$\epsilon$, produce high-quality results even for the more complex
three-class segmentation problem.  
Calculation times are also competitive
with those reported for the binary case (0.5s - 50s).

\subsection{Image Segmentation}

As another test setup, we use a grayscale image of size $191 \times
196$, taken from~\cite{jung:kang:shen,li:kim} and composed of 5 classes:
black, dark gray, medium gray, light gray and white. This image contains
structure, such as an internal hole and junctions where multiple
classes meet. The image information is represented through feature
vectors defined as $(x_i, y_i, \mathrm{pix}_i)$, with $x_i$ and $y_i$
corresponding to $(x, y)$ coordinates of the pixel and $\mathrm{pix}_i$
equal to the intensity of the pixel.  All of these are normalized so as
to obtain values in the range $[0,1]$.

\begin{figure*}[tb]
\centerline{
\subfigure{\scalebox{0.4}{\includegraphics{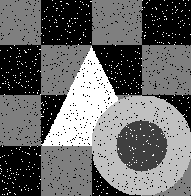}
\label{fig:img_fid}}}
\hfil
\subfigure{\scalebox{0.4}{\includegraphics{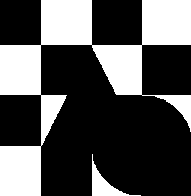}
\label{fig:class0}}}
\hfil
\subfigure{\scalebox{0.4}{\includegraphics{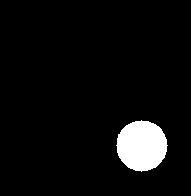}
\label{fig:class1}}}
}
\vspace{0.2cm}
\centerline{
\subfigure{\scalebox{0.4}{\includegraphics{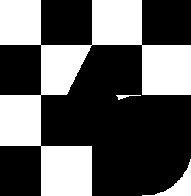}
\label{fig:class2}}}
\hfil
\subfigure{\scalebox{0.4}{\includegraphics{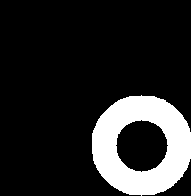}
\label{fig:class3}}}
\hfil
\subfigure{\scalebox{0.4}{\includegraphics{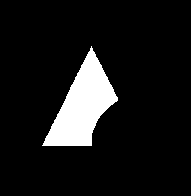}
\label{fig:class4}}}
}
\caption{Image Segmentation Results. Top left: Original five-class
image, with randomly chosen fidelity points displayed.  Other panels:
the five segmented classes, shown in white.}
\label{fig:image_segmentation}
\end{figure*}

The graph is constructed using $N = 30$ nearest neighbors and local
scaling based on the $M = 30$ closest point.  We use parameters
$\epsilon = 1$, $dt = 0.01$ and $m_{\mathrm{max}} = 800$.  We then choose 1500 random
points (4\% of the total)  for the fidelity term, with
$\lambda=30$. Figure~\ref{fig:image_segmentation} displays the original
image with the randomly selected fidelity points (top
left), and the five-class segmentation.  Each class image shows in white the
pixels identified as belonging to the class, and in black the pixels of
the other classes. In this case, all the classes are segmented
perfectly with an average run time of 59.7s. The method of Li and
Kim~\cite{li:kim} also segments this image perfectly, with a reported
run time of 0.625s.  However, their approach uses additional
information, including a pre-assignment of specific grayscale levels to
classes, and the overall densities of each class.  Our approach does not
require these.

\subsection{MNIST Data}

The MNIST data set available at
\textit{http://yann.lecun.com/exdb/mnist/} is composed of 70,000 images
of size $28 \times 28$, corresponding to a broad sample of handwritten
digits 0 through 9. We use the multiclass diffuse interface model to
segment the data set automatically into 10 classes, one per handwritten
digit.  Before constructing the graph, we preprocess the data by
normalizing and projecting into 50 principal components, following the
approach in~\cite{szlam:bresson}. No further
steps, such as smoothing convolutions, are required. The graph is computed
with $N = 10$ nearest neighbors and local scaling based on the $M =
10$ closest points. 

An adaptive $\epsilon$ variant of the algorithm is implemented, with
parameters $\epsilon_0 = 2$, $\epsilon_f = 0.01$, $\epsilon$ decrement
10\%, $dt = 0.01$, and 40 iterations per step. For the
fidelity term, 7,000 images (10\% of total) are chosen, with weight
$\lambda=30$. The average error
obtained, over 20 runs with four different
fidelity sets, is 7\% ($\pm 0.072\%$). The confusion matrix for the best
result obtained, corresponding to a 6.86\% error, is given in
Table~\ref{tab:mnist}: each row represents the segmentation obtained,
while the columns represent the true digit labels.  For reference,
the average computing time per run in this adaptive technique is 132s.
Note that, in the segmentations, the largest mistakes made are in trying
to distinguish digits 4 from 9 and 7 from 9.

For comparison, errors reported using unsupervised clustering algorithms in the
literature are: 12.9\% for p-Laplacian~\cite{buhler:hein}, 11.8\% for
ratio-minimization relaxed Cheeger cut~\cite{szlam:bresson}, and 12.36\% for
the multicut version of the normalized 1-cut~\cite{hein:setzer}.  A more
sophisticated graph-based diffusion method applied in a semi-supervised
setup (transductive classification), with function-adapted
eigenfunctions,
a graph constructed with 13 neighbors, and self-tuning with the 9th
neighbor reported in~\cite{szlam:maggioni:coifman} obtains an error of
7.4\%.  Results with similar errors are reported
in~\cite{liu:he:chang}. Thus, the performance of the multiclass
GL on this data set improves upon other published
results, while requiring less preprocessing
and a simpler regularization of the functions on the graph.

\begin{table*}[tb]
\caption{Confusion Matrix for the MNIST Data Segmentation.}
\label{tab:mnist}
\begin{center}
\begin{tabular}{|c|r|r|r|r|r|r|r|r|r|r|}
\hline
\multicolumn{1}{|c|}{\bf Obtained / True}  & \multicolumn{1}{|c|}{\bf 0} & \multicolumn{1}{|c|}{\bf 1} & \multicolumn{1}{|c|}{\bf 2} & \multicolumn{1}{|c|}{\bf 3} & \multicolumn{1}{|c|}{\bf 4} & \multicolumn{1}{|c|}{\bf 5} & \multicolumn{1}{|c|}{\bf 6} & \multicolumn{1}{|c|}{\bf 7} & \multicolumn{1}{|c|}{\bf 8} & \multicolumn{1}{|c|}{\bf 9} \\
\hline
{\bf 0} & 6712 & 3 &	39 &	10 &	6 & 36 & 57 & 10 & 61 & 28 \\	
\hline
{\bf 1} & 1	& 7738 &	7 & 15 & 9 & 1 & 9 & 	23 & 	36 &	12 \\	
\hline
{\bf 2} & 24 & 50 & 6632 &	 95 &  65 & 17 & 16 & 63 & 65 & 30 \\	
\hline
{\bf 3} & 13 &16	 & 84	 & 6585 & 8 & 218 & 5 & 42 & 153 & 84 \\
\hline
{\bf 4} & 5	& 6 & 27 & 8 & 6279	& 32	& 13	& 59	& 43	& 305 \\
\hline
{\bf 5} & 21 & 6	& 13	& 128 & 27 & 5736 & 57 & 3 & 262 & 34 \\	
\hline
{\bf 6} & 91 & 26 & 50 & 11 & 35 & 91 & 6693 & 0 & 45 & 1 \\	
\hline
{\bf 7} & 6	& 6 & 31 & 97 & 26 & 15 & 0 & 6689	& 24	& 331 \\	
\hline
{\bf 8} & 27 & 15 & 86 & 156 & 21 & 110 & 25 & 16 & 6065 & 66 \\	
\hline
{\bf 9} & 3	& 11	& 21	& 36	& 348 & 57 & 1	& 388 & 71 & 6067  \\
\hline
\end{tabular}
\end{center}
\end{table*}

\section{\uppercase{Conclusions}}
\label{sec:conclusion}

\noindent We have proposed a new multiclass segmentation procedure, based on the
diffuse interface model. The method obtains segmentations of
several classes simultaneously without using one-vs-all or alternative
sequences of binary segmentations required by other multiclass methods.
The local scaling method of Zelnik-Manor and Perona, used to construct
the graph, constitutes a useful representation of the characteristics of
the data set and is adequate to deal with high-dimensional data. 

Our modified diffusion method, represented by the non-linear smoothing
term introduced in the Ginzburg-Landau functional, exploits the
structure of the multiclass model and is not affected by the ordering of
class labels.  It efficiently propagates class information that is known
beforehand, as evidenced by the small proportion of fidelity points (4\%
- 10\% of dataset) needed to perform accurate segmentations.  Moreover, the
method is robust to initial conditions.  As long as the initialization
represents all classes uniformly, different initial random configurations
produce very similar results.  The main limitation of the method appears
to be that fidelity points must be representative of class distribution.
As long as this holds, such as in the examples discussed, the
long-time behavior of the solution relies less on choosing the ``right''
initial conditions than do other learning techniques on graphs.

State-of-the-art results with small classification errors were obtained
for all classification tasks. Furthermore, the results do not depend on
the particular class label assignments. Future work includes
investigating the diffuse interface parameter $\epsilon$.  We conjecture
that the proposed functional converges (in the $\Gamma$-convergence
sense) to a total variational type functional on graphs as $\epsilon$ approaches zero, but the exact nature of the limiting functional is unknown.

\section*{\uppercase{Acknowledgements}}
This research has been supported by the Air Force Office of Scientific Research MURI grant FA9550-10-1-0569 and by ONR grant  N0001411AF00002.

\bibliographystyle{apalike}
{\small
\bibliography{ref_graph}}


\vfill
\end{document}